\documentclass[lettersize,journal]{IEEEtran}
\usepackage{amsmath,amsfonts}
\usepackage{algorithmic}
\usepackage{algorithm}
\usepackage{array}
\usepackage[caption=false,font=normalsize,labelfont=sf,textfont=sf]{subfig}
\usepackage{textcomp}
\usepackage{stfloats}
\usepackage{url}
\usepackage{bm}
\usepackage{verbatim}
\usepackage{graphicx}
\usepackage{cite}

\usepackage{xcolor}
\usepackage{booktabs}
\usepackage{arydshln}
\usepackage{float}
\usepackage{longtable}
\usepackage{graphicx}
\usepackage{pdflscape}
\usepackage{amssymb}
\usepackage{siunitx}
\usepackage{hyperref}

\usepackage{pgfplots}
\pgfplotsset{compat=1.17}

\begin{document}

\title{Long-Tailed Continual Learning For \\ Visual Food Recognition}

\author{Jiangpeng He,~\IEEEmembership{Member,~IEEE,} Xiaoyan Zhang, Luotao Lin, Jack Ma, \\ Heather A. Eicher-Miller, and Fengqing Zhu,~\IEEEmembership{Senior Member,~IEEE}

        }

%

\markboth{Journal of \LaTeX\ Class Files,~Vol.~14, No.~8, August~2021}%
{Shell \MakeLowercase{\textit{et al.}}: A Sample Article Using IEEEtran.cls for IEEE Journals}


\maketitle


\begin{abstract}
Deep learning-based food recognition has made significant progress in predicting food types from eating occasion images. However, two key challenges hinder real-world deployment: (1) continuously learning new food classes without forgetting previously learned ones, and (2) handling the long-tailed distribution of food images, where a few common classes and many more rare classes. To address these, food recognition methods should focus on long-tailed continual learning. In this work, We introduce a dataset that encompasses 186 American foods along with comprehensive annotations. We also introduce three new benchmark datasets, VFN186-LT, VFN186-INSULIN and VFN186-T2D, which reflect real-world food consumption for healthy populations, insulin takers and individuals with type 2 diabetes without taking insulin. We propose a novel end-to-end framework that improves the generalization ability for instance-rare food classes using a knowledge distillation-based predictor to avoid misalignment of representation during continual learning. Additionally, we introduce an augmentation technique by integrating class-activation-map (CAM) and CutMix to improve generalization on instance-rare food classes. Our method, evaluated on Food101-LT, VFN-LT, VFN186-LT, VFN186-INSULIN, and VFN186-T2DM, shows significant improvements over existing methods. An ablation study highlights further performance enhancements, demonstrating its potential for real-world food recognition applications.

\end{abstract}

\vspace{-0.25cm}

\section{Introduction}
\label{sec: inctroduction}
The emergence of modern deep learning technologies has enabled automatic food nutrition analysis, including image-based dietary assessment~\cite{boushey2017new, zhu2010use, he2021end, he2020multitask}, to monitor and improve dietary intake and prevent chronic diseases like diabetes. As the first step in this process, food recognition identifies food types from images, and accurate recognition is critical for overall assessment performance. Deep learning-based methods~\cite{min2019survey, Qiu2019MiningDF, mao2020visual, IBM} demonstrate remarkable performance by training off-the-shelf Convolutional Neural Networks (\textit{e.g.} ResNet\cite{RESNET}) using static datasets (\textit{e.g.} Food-101\cite{Food-101}, Food2K~\cite{Food2K}). However, two major challenges remain in real-world applications: (i) updating models as new food classes emerge over time, and (ii) addressing the severe class imbalance in long-tailed distributions, where a few classes (head classes) dominate consumption compared with most others (tail classes)~\cite{he2022long, he2023single}. Failing to address these can significantly degrade performance.

Continual learning, also known as incremental or lifelong learning, allows models to learn new classes continuously without catastrophic forgetting~\cite{CF,lu2024visualprompttuningnull,lu2025training,hu2024taskaware,cheng2025achieving}. Unlike retraining from scratch whenever encountering a new class, continual learning is more practical, requiring only new class data, which improves time, computation, and memory efficiency~\cite{JH_THESIS}. The challenge intensifies when the data follows a long-tailed distribution~\cite{raghavan2024online, 10180221}, requiring the model to address both catastrophic forgetting and class imbalance. While recent work~\cite{liu2022long} introduces a 2-stage framework to tackle this, its manual fine-tuning and detached training stages pose inefficiencies for real-world use. Additionally, existing methods have not been specifically applied to food images. This presents further challenges because food images often exhibit high intra-class variation and inter-class similarity, making it difficult to distinguish between different food items.

Existing continual learning methods show the effectiveness of applying knowledge distillation and storing a small fixed number of seen images as exemplars to mitigate catastrophic forgetting. However, both techniques become less effective in the long-tailed distribution. Knowledge distillation~\cite{KD},\cite{10473168} may even harm the performance when the teacher's model is not trained on balanced data due to the bias in output logits as shown in a recent study~\cite{dualmemory}. On the other hand, distilling knowledge through learned representations imposes a new challenge of feature space misalignment~\cite{wang2021distilling} as the learned representation needs to evolve during continual learning to accommodate new classes. Regarding using an exemplar set, most classes in long-tailed distribution may contain only a few training samples. Consequently, the overall performance may still be hindered even when all available samples are stored for instance-rare classes due to the poor generalization ability. 

In this work, we focus on designing an end-to-end long-tailed continual learning framework for visual food recognition. We leverage feature-based knowledge distillation while incorporating an additional prediction head that projects the current representation space to the past. This addresses the misalignment issue by providing more freedom to the student model and encourages the retention of the learned knowledge. In addition, inspired by the most recent work~\cite{CMO} that uses the context-rich information in head classes to help the tail classes, we introduce a new data augmentation technique by integrating class-activation-map (CAM) and CutMix~\cite{yun2019cutmix}, which cuts the most important region calculated by CAM in instance-rare classes data as foreground and pastes into the instance-rich classes images. With minimal computational overhead, this method significantly enhances the generalization capabilities of tail classes. We evaluate our method on existing long-tailed food image datasets including Food101-LT and VFN-LT~\cite{he2022long}. Additionally, we developed VFN186 based on the original VFN~\cite{mao2020visual}, expanding the initial 74 food categories by adding 112 more. This allows for a more comprehensive coverage of the typical American diet~\cite{wweia2015}. Furthermore, we derive three long-tailed versions of VFN186, referred to as VFN186-LT, VFN186-INSULIN, and VFN186-T2D, based on different population groups, namely healthy populations, Insulin Takers, and those with Type 2 diabetes without taking insulin. Our proposed framework achieves the best performance with a large improvement margin compared to existing methods while not requiring detached training stages. Finally, we conduct an ablation study to evaluate the effectiveness of each component in our method and discuss potential techniques that can boost the accuracy for real-world applications. The main contributions of this work are summarized in the following: 

\begin{itemize}  
    \item We introduce the VFN186 food dataset, which contains 186 most frequently consumed food types in America. Additionally, we introduce three new long-tailed benchmark datasets, which reflect the food consumption patterns of different populations. The dataset will be public. 
    \item We propose a novel framework that utilizes feature-based knowledge distillation with a prediction head and a novel CAM-based CutMix for data augmentation, and an integrated loss function to address catastrophic forgetting and class imbalance.
    \item We conduct extensive experiments on all long-tailed continual learning benchmarks for food recognition and discuss potential techniques to enhance accuracy that could boost the accuracy for facilitating the deployment in real-world food recognition.
\end{itemize}

\vspace{-0.5em}
\section{Related Work}
\label{sec: related_worl}
In this section, we summarize existing methods most related to our work including food recognition, long-tailed recognition, and continual learning. 

\vspace{-1em}

\subsection{Food Recognition}
\label{subsec: food recognition}
 Food image recognition is a challenging yet practical task with applications like image-based dietary assessment~\cite{kitamura2008food, shao2021_ibdasystem}, where accurate recognition is crucial for nutritional content analysis, such as energy and macronutrients~\cite{fang2019end, vinod2024food, ma2024mfp3d, shao2023end}. Most existing deep learning based work leverage off-the-shelf models~\cite{googlenet, vgg, RESNET, densenet} and train on static food image datasets~\cite{Food-101, Food2K, uec-256, upmc, vireo172, mao2020visual}. To address inter-class similarity and intra-class variability, various hierarchy-based approaches have been proposed~\cite{IBM, mao2020visual, pan2023muti}. While food recognition has been studied in scenarios like ingredient recognition~\cite{chen2020study}, fine-grained recognition~\cite{rodenas2022learning, pan2024fmifood}, few-shot learning~\cite{jiang2020few}, long-tailed recognition~\cite{he2022long, gao2022dynamic_LTingredient, he2023single}, and continual learning~\cite{He_2021_ICCVW, raghavan2024online}, no existing methods continuously learn new classes in long-tailed distributions, which is critical for real-world applications~\cite{he2022long}. Recent work~\cite{liu2022long} attempted to integrate continual learning with long-tailed recognition but used a multi-stage training process and did not focus on food images. In this work, we target long-tailed continual learning for visual food recognition, introducing a novel end-to-end framework to address both class imbalance and catastrophic forgetting simultaneously.

\vspace{-1.2em}

\subsection{Long-tailed Recognition}
\label{subsec: Long-tailed Recognition}
 Existing work on image recognition in long-tailed distributions can be categorized into two main groups: \textit{re-weighting} and \textit{re-sampling}. The major challenge is the imbalance between instance-rich (head) and instance-rare (tail) classes~\cite{Cheng_2022}. \textbf{Re-weighting} methods balance the loss or gradients during training, with a class-level re-weighting loss like Balanced Softmax~\cite{BSLoss} and Label-Distribution-Aware Margin loss~\cite{LDAM}.
 In addition, \textbf{re-sampling} based techniques construct a balanced training set by over-sampling tail classes or under-sampling head classes, but naive over-sampling~\cite{ROS} and  under-sampling~\cite{RUS_ROS} can lead to overfitting or performance degradation. 
 Therefore, most existing work performs data augmentation to improve the generalization ability of tail classes and achieve better overall performance. Gao \textit{et al}.~\cite{gao2022dynamic_LTingredient} propose Dynamic Mixup for multi-label long-tailed recognition problem, which dynamically adjusts the selection of images based on the previous training performance and set the label of synthetic image as the union of two images. CMO~\cite{CMO} applies CutMix~\cite{yun2019cutmix} for data augmentation by cutting the foreground region in tail classes images and pasting in head classes background. The center idea of CMO is to leverage the context rich information from the head classes to help the generalization of tail classes. Later, He \textit{et al}.~\cite{he2022long} improves the CMO to use visually similar image pairs and allows for multi-image CutMix to achieve improved performance. In spite of the efficiency of CutMix for data augmentation, one of the limitations is that it suffers from loss of semantic information of the original image since the cut region is generated randomly. Inspired by~\cite{huang2021snapmix}, we introduce a novel CAM-based CutMix, which combines the images seamlessly without losing semantic information, as detailed in Section~\ref{sec: method}.

\vspace{-1.0em}

\subsection{Continual Learning}
\label{subsec: continual learning}

Continual learning, also known as incremental or lifelong learning, has been explored in scenarios like class-incremental, task-incremental, and domain-incremental learning~\cite{hsu2018re}. This work focuses on class-incremental learning, which is key for real-world applications. It involves continuously learning new classes and classifying all previously seen classes during inference, without using task indexes or multi-head classifiers as in task-incremental learning~\cite{SIT}. Unlike domain-incremental learning, which handles domain shifts without new classes, class-incremental learning faces the challenge of catastrophic forgetting~\cite{CF}, where the model forgets previous knowledge due to the lack of data from learned classes~\cite{cheng2024efficient}. To address this, existing methods are mainly divided into \textit{regularization-based} and \textit{replay-based} approaches.

\textbf{Regularization-based} methods address forgetting by limiting changes to learned parameters while learning new classes. Initial work froze or constrained parameter updates~\cite{jung2016less, kirkpatrick2017overcoming}, limiting the model's ability to learn new data. Later, Li \textit{et al}.~\cite{LWF} used knowledge distillation\cite{KD} to preserve learned knowledge by mimicking the teacher model's output logits. 
Feature-based distillation minimized representation discrepancies~\cite{rebalancing} of learned representations between student and teacher models, which is further developed in~\cite{douillard2020podnet} integrating the logits and feature-based distillation. However, the knowledge distillation using output logits may even harm the overall performance if the teacher model is not trained on balanced data due to the bias towards instance-rich classes.  Furthermore, direct feature-based distillation also faces challenges like feature space misalignment~\cite{wang2021distilling} due to the evolving of feature space when learning new classes especially in a long-tailed scenario where the data distribution may vary a lot for each incremental learning step.
We address this problem by adding a prediction head to map the current representation space to the past, enabling more efficient knowledge transfer.

\textbf{Replay-based} methods use a memory buffer to store exemplar data for knowledge replay during class-incremental learning. The herding algorithm~\cite{HERDING} selects exemplars based on class mean vectors and is widely used~\cite{ICARL, ILIO, EEIL, BiC}. However, these methods assume balanced training data and sufficient samples per class compared with the memory budget (\textit{e.g. }20 exemplars per class), which isn't the case in long-tailed scenarios. It can lead to class imbalance in the exemplar set and harm overall performance. We address this issue by constructing a balanced exemplar set by augmenting the tail class data with the proposed CAM-based CutMix, which augments tail class data, improving knowledge replay efficiency and generalization on tail classes for better overall performance.

\vspace{-1.2em}

\subsection{Continual Learning with Pre-trained Models}
\label{subsec: pretrained CL}
Leveraging large-scale pre-trained vision transformers for continual learning has emerged as a promising direction.
Recent works optimize prompt parameters, a small set of learnable weights, to guide model predictions without storing past examples, which enable efficient knowledge encoding and transfer across tasks.
L2P~\cite{DBLP:journals/corr/abs-2112-08654} pioneers prompt-based continual learning, which encodes knowledge through learnable prompt parameters to facilitate efficient adaptation. 
Wang \textit{et al} proposes DualPrompt~\cite{wang2022dualpromptcomplementarypromptingrehearsalfree}, which learn two sets of disjoint prompt spaces to encode task-invariant and task-specific instructions respectively. CODA-Prompt~\cite{smith2023codapromptcontinualdecomposedattentionbased} further refines them using decomposed prompts, which dynamically assemble into attention-conditioned prompts optimized in an end-to-end manner. These methods leverage pre-trained Vision Transformer (ViT) and demonstrate strong performance on benchmark datasets, highlighting the potential of prompt-based approaches in continual learning~\cite{smith2023codapromptcontinualdecomposedattentionbased}.

\vspace{-0.5em}
\section{Datasets}
\label{sec: dataset}
\vspace{-0.5em}
\subsection{VFN186 Dataset}
To encompass a broader range of food types, we expanded VFN~\cite{mao2020visual} to include 186 food categories\footnote{The dataset is available at \url{https://github.com/JiangpengHe/VFN186}}. Similar to VFN, we select an additional 112 commonly consumed foods by Americans based on What We Eat In America (WWEIA)~\cite{wweia2015} database. This expansion makes the dataset more comprehensive and robust for training practical models with broader applicability. Specifically, similar to~\cite {he2022long, chen2024metafood3d}, we first match each of the 186 food types in VFN186 with one 8-digit USDA food code from the Food and Nutrient Database for Dietary Studies (FNDDS) where each 8-digit USDA food code represents a specific food item in the food supply. Then we use a semi-automatic data collection system to crawl specific types of food images from the Google Image website based on food labels. Next, we employ a trained Faster R-CNN~\cite{ren2016faster} to remove noisy images. The remaining images were processed using an online crowdsourcing tool, where food items were boxed and labeled with their corresponding categories. Through this process, we expand the VFN dataset and created the VFN186 dataset, which includes 186 food types and 70230 images.

\vspace{-1em}
\subsection{Long-tailed Food Datasets For Different Populations}
While our VFN186 dataset provides rich value for various downstream tasks, in this work, we primarily focus on its long-tailed version for continual learning, which leverages its strength of matched food codes from the nutrition database and addresses the challenges of real-world food data distribution across different populations. Specifically, we generated three long-tailed versions of our VFN186 dataset. First, \textbf{VFN186-LT} follows the methodology of~\cite{he2022long}, reflecting real-world food consumption frequencies among \textit{healthy} individuals aged 18 to 65 in U.S. as reported by~\cite{lin2022differences}. 

Additionally, we developed \textbf{VFN186-INSULIN} and \textbf{VFN186-T2D}, designed for dietary assessment among \textit{insulin takers} and those with \textit{type 2 diabetes without taking insulin}, respectively. The motivation of developing VFN186-INSULIN and VFN186-T2D lies in addressing critical gap in food recognition research for the approximately 34.2 million U.S. individuals (10.5\% of the population) with diabetes~\cite{diabete_2020, lin2024diet}. Given the crucial role of diet in managing diabetes and its associated health complications, these population-specific long-tailed datasets aim to enhance the practical applicability of food recognition models in real-world scenarios.

\begin{figure*}[t]
\begin{center}
  \includegraphics[width=1.\linewidth]{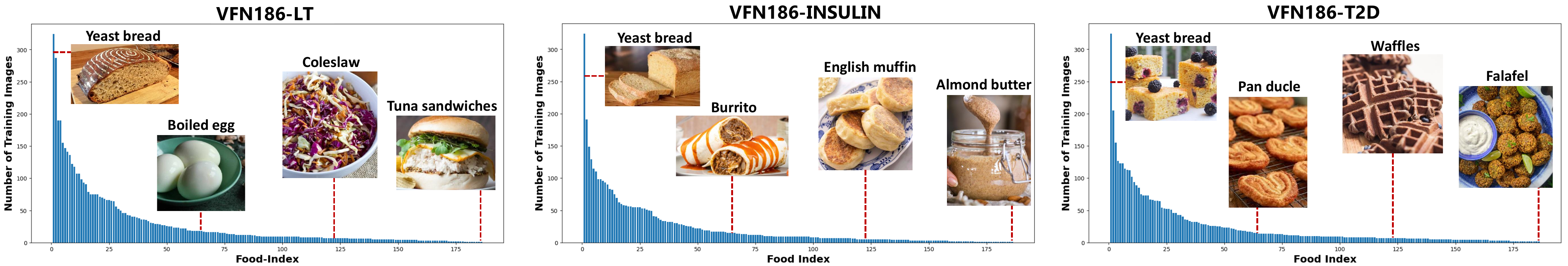}
  \vspace{-1em}
  \caption{The distribution of VFN186-LT, VFN186-INSULIN, and VFN186-T2D shown in descending order based on the number of training samples.}
  \label{fig:dataset}
\end{center}
\vspace{-1em}
\end{figure*}

The process of generating the long-tailed datasets follows ~\cite{he2022long} to reduce the number of training samples for food classes in the original VFN186 based on the matched consumption frequency. Overall, VFN186-LT contains 5,185 training images across 186 classes, VFN186-INSULIN contains 4,179 training images, and VFN186-T2D contains 4,403 training images, with a maximum of 324 and a minimum of 1 image per class. The imbalance ratio $\rho$, defined as the maximum over the minimum number of training samples, equals $324$ for three datasets.
The food type \textit{Yeast bread} is highly consumed among represented adults and dominates the consumption frequency in all groups, while frequencies of other food types vary between them. 
Figure~\ref{fig:dataset} shows the distribution of food types in VFN186-LT, VFN186-INSULIN, and VFN186-T2D, ranked by the number of training samples per class.

\begin{figure*}[t]
\begin{center}
  \includegraphics[width=1.\linewidth]{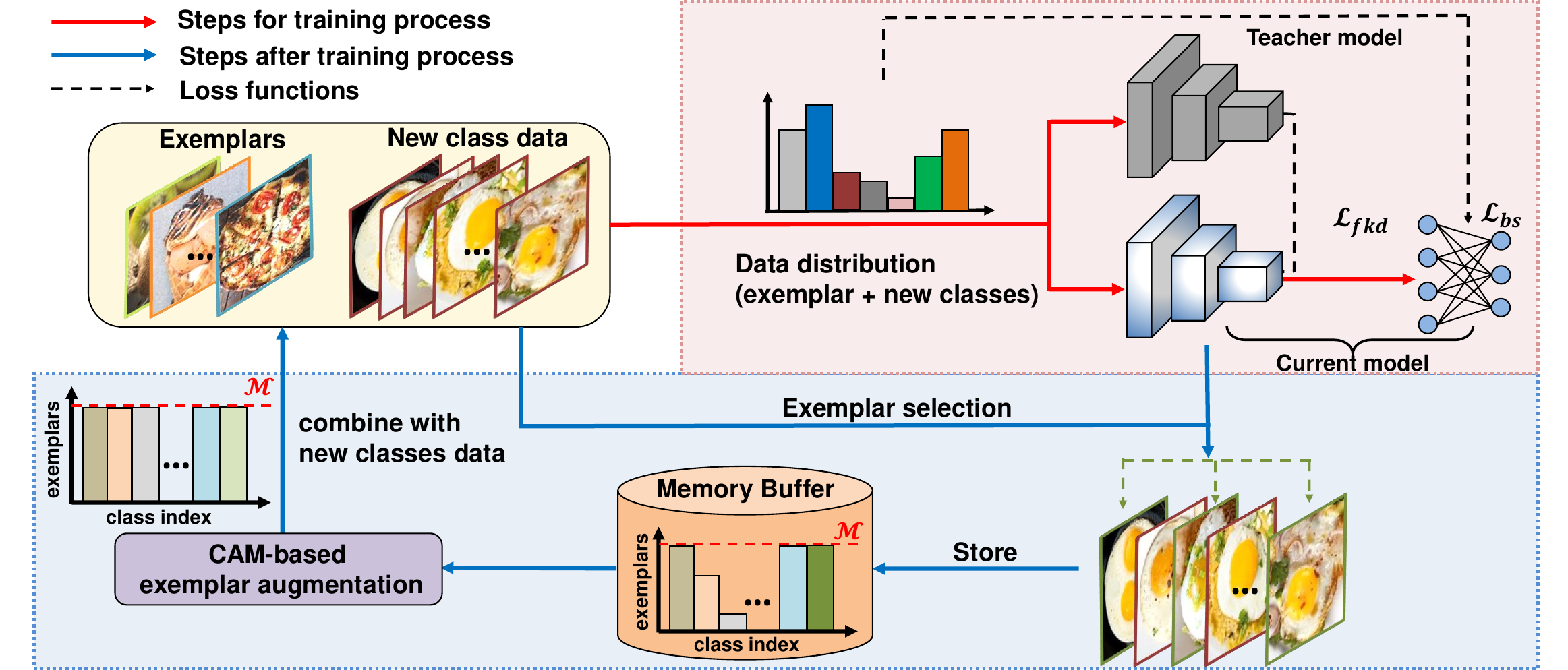}
  \caption{The overview of our proposed framework. The red arrows show the training process with new class images and exemplars from previous classes. The blue arrows denote the steps after the training process where we construct a balanced exemplar set and store them in the memory buffer. }
  \label{fig:method}
\end{center}
\vspace{-1.5em}
\end{figure*}

\vspace{-0.5em}
\section{Method}
\label{sec: method}
In this work, we introduce a novel end-to-end long-tailed continual learning framework for visual food recognition. The overview of our method is shown in Figure~\ref{fig:method}. To address catastrophic forgetting, we leverage the teacher model learned from the last incremental step and perform feature-based knowledge distillation with an additional prediction head to enable efficient knowledge transfer. The exemplar set was selected based on a novel CAM-based data augmentation for tail classes. Finally, we replace the cross-entropy with the balanced softmax loss~\cite{BSLoss} based on the current training data distribution to learn class-balance visual representation. In this section, we first introduce the preliminaries for continual learning in the long-tailed distribution in Section~\ref{subsec: preliminaries} and then illustrate the detail of each proposed component in Section~\ref{subsec: Feature-based Knowledge Distillation},~\ref{subsec: CAM-based Image Augmentation} and \ref{subsec: loss}, respectively. 

\vspace{-0.5em}

\subsection{Preliminaries}
\label{subsec: preliminaries}
We focus on continual learning in class-incremental settings where the objective is to learn new classes incrementally and perform classification on all classes seen so far during the inference phase. Specifically, the continual learning in the class-incremental scenario can be formulated as applying an initial model $h_0$ to learn a sequence of $N$ tasks denoted as $\mathcal{T} = \{\mathcal{T}^1,...,\mathcal{T}^N\}$ where each task $\mathcal{T}^i$ contains $C_i$ non-overlapped new classes, which is also known as the incremental step size. During the learning phase of each new task, only the training data $D_i = \{\textbf{x}_i^j, y_i^j\}$ of the current task is available where $\textbf{x}_i^j$ and $y_i^j$ denote the $j$-th input image and label, respectively. After each incremental learning step, the updated model $h_i$ needs to classify $C_{1:i}$ classes encountered so far. The major challenge of continual learning is catastrophic forgetting~\cite{CF} where the updated model $h_i$ after learning the task $\mathcal{T}^i$ forgets the knowledge of previous tasks $\{\mathcal{T}^1,...,\mathcal{T}^{i-1}\}$, resulting in significant performance degradation to classify $C_{1:i-1}$. In the conventional setup, the training data $D_i$ for each task $\mathcal{T}^i$ is evenly distributed, containing $\vert D_i \vert / C_i$ samples per class. However, this assumption simplifies the real-world complexities, especially for food recognition where data is usually long-tailed distributed and exhibits imbalance among food classes. Formally, the training data $D_i$ for each task in long-tailed continual learning is a class-imbalanced distribution with each class containing $(0, \vert D_i \vert)$ training samples. The entire training data $D$ for all the $N$ tasks $\mathcal{T}$ exhibits the long-tail distribution. 

\subsubsection{Knowledge distillation}
\label{subsubsection: kd}
Most existing work~\cite{LWF, BiC, ICARL, EEIL, ILIO} applies knowledge distillation~\cite{KD} on output logits to maintain the performance on previously learned classes. Specifically, during the learning step of the task $\mathcal{T}^i$, a teacher model $h_t = h_{i-1}$ learned from the last task with fixed parameters is employed. The knowledge distillation aims to minimize the difference between the output logits of the current model $L = [o^1, o^2, ... o^{C_{1:i}}] \in   \mathbb{R}^{C_{1:i} \times 1}$ and the outputs of the teacher model $\hat{L} = [\hat{o}^1, \hat{o}^2, ... \hat{o}^{C_{1:i-1}}] \in \mathbb{R}^{C_{1:i-1} \times 1}$ by 
\begin{equation}
\label{eq: old kd}
    \mathcal{L}_{kd} = -\sum_{j=1}^{C_{1:i-1}}\hat{L}^{(j)}_T log(L^{(j)}_T)
\end{equation}
where T is the temperature scalar to learn the hidden knowledge by softening the output distribution as
\begin{equation}
\label{eq: temp T}
    \hat{L}_T^{(j)} =
\frac{\exp{(\hat{o}^{(j)}}/T)}{\sum_{k=1}^{C_{1:i-1}}\exp{(\hat{o}^{(k)}}/T)}
\end{equation}
Finally, the knowledge distillation is integrated with cross-entropy during the training process by using a hyper-parameter $\alpha$ to learn new classes and maintain the learned knowledge. 
\begin{equation}
    \label{eq: og_crosskd}
    \mathcal{L} = \alpha \mathcal{L}_{kd} + (1-\alpha) \mathcal{L}_{cn}
\end{equation}

\subsubsection{Exemplar replay}
\label{subsubsection: exemplar}
As one of the most commonly used strategies to address catastrophic forgetting, the exemplar replay-based methods~\cite{ICARL, EEIL, rebalancing} assume the availability of a reasonable memory budget to select a small fixed number of data as exemplars for each seen class and store them in memory buffer (also known as exemplar set). Specifically, after learning each task $\mathcal{T}^i$, the lower layers of updated model $h_i$ are used to extract feature embeddings for the new classes training data $D_i = \{\textbf{x}_i^j, y_i^j\}$. The Herding algorithm~\cite{HERDING} is widely applied to select the most representative data for each new class based on the Euclidean distance between feature embedding and the class mean vector. Therefore, given a memory budget of $\mathcal{M}$ data per class (also known as memory capacity), a subset of $E_i \subseteq D_i$ is selected with $\vert E_i \vert = \mathcal{M} \times C_i$ and stored in the memory buffer. Finally, at the beginning of the next new task $\mathcal{T}^{i+1}$, all the exemplars in the memory buffer are combined with the new classes training data to construct $E_i + D_{i+1}$ for continual learning. In this work, we use Herding as the exemplar selection algorithm while other latest work~\cite{He_2021_ICCVW} could also be applied. 

\subsection{Feature-based Knowledge Distillation}
\label{subsec: Feature-based Knowledge Distillation}
\begin{figure}[t]
\begin{center}
  \includegraphics[width=1.\linewidth]{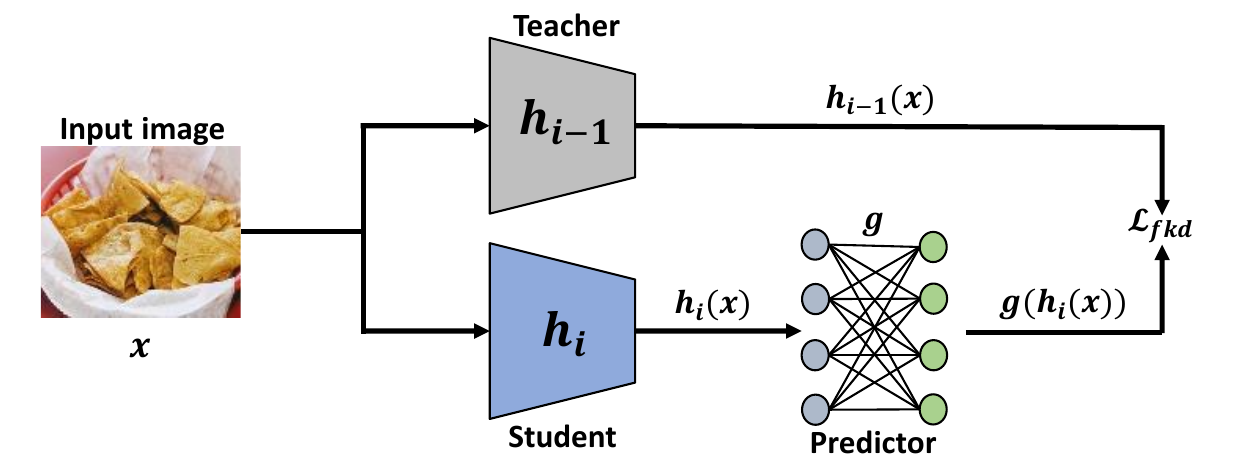}
  \caption{The overview of proposed feature-based knowledge distillation by applying an additional predictor $g$. }
  \label{fig:method-fkd}
\end{center}
\vspace{-0.7cm}
\end{figure}

Despite the effectiveness of knowledge distillation in conventional continual learning setup as described in Section~\ref{subsubsection: kd}, it is difficult to apply it to long-tailed distributions since the output logits of the teacher model can be heavily biased towards instance-rich classes~\cite{BiC}. Directly applying knowledge distillation as in (\ref{eq: old kd}) on biased output logits may even harm the overall performance~\cite{dualmemory}. Therefore, we explore feature-based knowledge distillation for better knowledge transfer in long-tailed continual learning. However, a key challenge is feature space misalignment of the challenges when applying feature-based distillation, where the representation of student and teacher models could mismatch in terms of both magnitude and direction~\cite{wang2021distilling}. This problem is also relevant in continual learning as the model evolves to incorporate new classes. To solve this, we introduce a simple yet effective method as shown in Figure~\ref{fig:method-fkd}. Specifically, instead of directly mimicking the feature from the teacher model, we apply an additional predictor $g$ on the head of the continual learning model to map the current representation space to the past in the teacher model. 
$g$ is a single-layer perceptron that performs domain mapping while preserving consistent dimensions. Specifically, it maps from $\mathbb{R}^{d \times 1}$ to $\mathbb{R}^{d \times 1}$, followed by a ReLU activation function. The dimensional consistency is crucial to ensure that the student model, with the added $g$, still outputs image features of the same size, i.e., $d \times 1$.
Given an image $\textbf{x}$, the predictor $g$ takes the feature representation from the current model (\textit{i.e. }student model) $h_i(\textbf{x})$ as input and outputs the mapped feature $g(h_i(\textbf{x}))$. Then we distill the knowledge from the teacher model $h_{i-1}$ by 
\begin{equation}
    \label{eq: fkd}
    \mathcal{L}_{fkd}(\textbf{x})) = 1 - <g(h_i(\textbf{x})), h_{i-1}(\textbf{x}))>
\end{equation}
where $< , > $ measures the cosine similarity. By applying the predictor, we give the student model more freedom to accommodate the previously learned representation into the current feature space, enabling more efficient knowledge distillation in long-tailed continual learning. The predictor $g$ is removed after each incremental learning phase. Note that although we apply cosine embedding loss for knowledge distillation, our method can be integrated with other loss functions such as the Mean Squared Error (MSE) loss. 

\subsection{CAM-based Exemplar Augmentation}
\label{subsec: CAM-based Image Augmentation}
\begin{figure}[t]
\begin{center}
  \includegraphics[width=1.\linewidth]{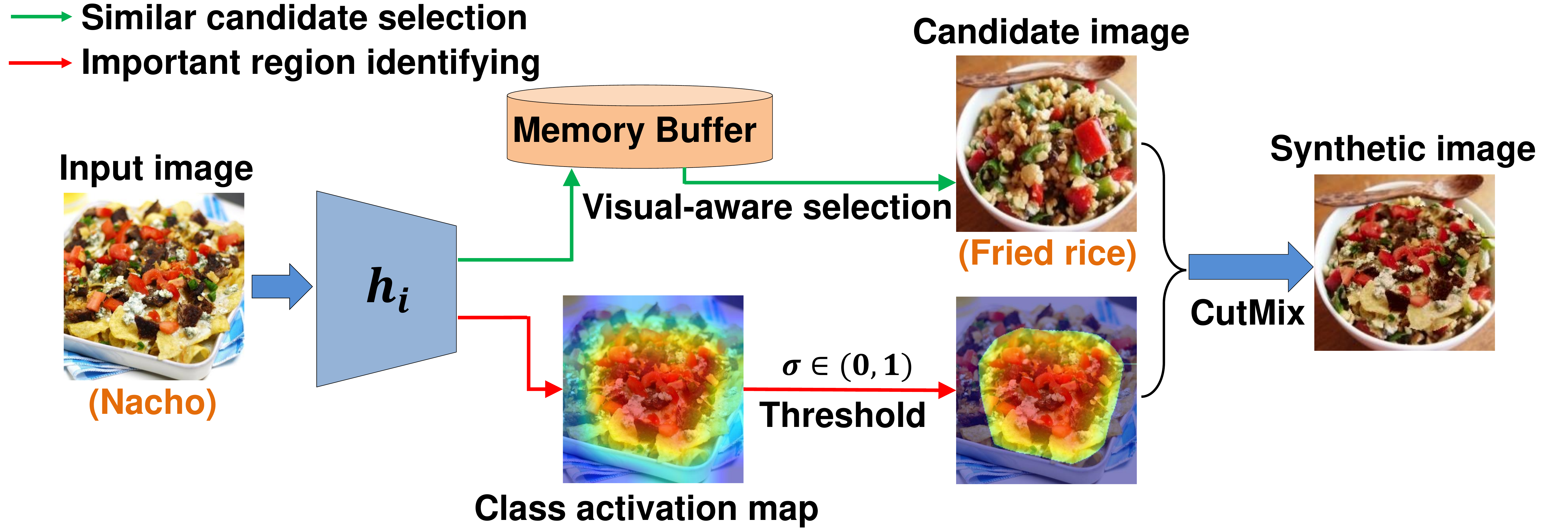}
  \caption{The overview of proposed CAM-based data augmentation technique. The green arrow describes the selection of the most visually similar candidate image and the red arrow illustrates the steps to obtain the most important region of the input image to perform CutMix~\cite{yun2019cutmix}.}
  \label{fig:cam}
\end{center}
\vspace{-0.7cm}
\end{figure}

Existing exemplar replay-based methods assume each class should contain at least $\mathcal{M}$ images given $\mathcal{M}$ as the memory budget in Section~\ref{subsubsection: exemplar}. However, most classes in long-tailed distribution may contain only a few training samples $n < \mathcal{M}$, which imposes two new challenges including (i) inefficiency of knowledge replay due to the insufficient training samples and (ii) intensification of the class-imbalance issue if we directly combine the stored exemplars with training data from new class due to the imbalanced nature of memory buffer. Therefore, we propose a novel data augmentation method in this work to construct a balanced exemplar set by augmenting the tail class images to address both aforementioned issues. The overview of the proposed data augmentation technique is illustrated in Figure~\ref{fig:cam}. To address the issue of losing semantic information when performing data augmentation~\cite{CMO, he2022long} as described in Section~\ref{subsec: Long-tailed Recognition}, we propose to use a class activation map (CAM)~\cite{Zhou_2016_CVPR_CAM} to identify the most important region from instance-rare classes images and then preserve the semantic information by performing CutMix~\cite{yun2019cutmix} to cut and paste the identified region into the images with rich context that are selected based on visual similarity. Specifically, we construct a class-balanced memory buffer before each new task $\mathcal{T}^{i+1}$ by augmenting stored images for food classes $C_t$ with less than $\mathcal{M}$ exemplars through CutMix in conjunction with images selected from food classes $C_h$ containing $\mathcal{M}$ exemplars. Given an input image $\textbf{x}_t \in C_t$, we first select the most visually similar candidate $\textbf{x}_h \in C_h$ by comparing the cosine similarity with $h_i$ as feature extractor where $\textbf{x}_h = \underset{\textbf{x}_k \in C_h}{\text{argmax}} <h_i(\textbf{x}_t), h_i(\textbf{x}_k)>$. The lower half of Figure~\ref{fig:cam} illustrates the procedure to identify the region to cut and paste into $\textbf{x}_h$. Formally, given $\textbf{x}_t \in \mathbb{R}^{c\times h \times w}$, the class-activation map $M(\textbf{x}_t) \in \mathbb{R}^{h \times w}$ is calculated by 
\begin{equation}
    \label{eq: cam}
    M(\textbf{x}_t) = \sum_k^dv_{y_{\textbf{x}_t}}^kh_i(\textbf{x}_t)
\end{equation}
where $v_{y_{\textbf{x}_t}} \in \mathbb{R}^d$ refers to the weight vector in the classifier of the current model corresponding to the seen class $y_{\textbf{x}_t} \in C_{1:i}$. The value of CAM ranges from $[0, 1]$ and a higher value indicates the more discriminative class-specific region. Therefore, we apply a random threshold $\sigma \in (0, 1) $ to select the region $M(\textbf{x}_t)^T \in \mathbb{R}^{h \times w}$ where 
\begin{equation} \label{eq:threshold}
M(\textbf{x}_t)^T=\left\{
\begin{array}{cl}
M(\textbf{x}_t) &  {M(\textbf{x}_t) \geq \sigma}\\0 & {M(\textbf{x}_t) < \sigma}
\end{array} \right.
\end{equation}
without losing the semantic information of the input image. Finally, we apply CutMix to generate a synthetic image $\tilde{\textbf{x}}_t$ by 
\begin{equation}
    \label{eq: cutmix}
    \tilde{\textbf{x}}_t = (1-S(\textbf{x}_t)^T) \odot \textbf{x}_h + S(\textbf{x}_t)^T \odot \textbf{x}_t 
\end{equation}
where $\odot$ refers to element-wise multiplication and $S(\textbf{x}_t)^T$ denotes the binary mask obtained from $M(\textbf{x}_t)^T$ that $\textbf{1}$ indicates the region with $M(\textbf{x}_t)^T > 0$. The class label $\tilde{y}_t$ of the synthetic image is calculated by the area of the replaced region in $\textbf{x}_h$ as 
\begin{equation}
    \label{eq: class-label}
    \tilde{y}_t = \frac{1 - A_r}{A}y_h + \frac{A_r}{A}y_t 
\end{equation}
where $A_r$ and $A$ denote the area of the replaced region and the total area of $\textbf{x}_h$, and $y_h$ and $y_t$ are the original class labels of $\textbf{x}_h$ and $\textbf{x}_t$. The exemplar augmentation is performed at the beginning of each new task and the augmented images are not stored in the memory buffer. Note that the Grad-CAM\cite{selvaraju2017gradcam}, which can be regarded as the generalization of CAM~\cite{Zhou_2016_CVPR_CAM}, could also be applied in our method. 

\vspace{-1em}
\subsection{Integrated Loss}
\label{subsec: loss}
While the exemplar augmentation mitigates the class-imbalance issue by constructing a balanced memory buffer, the number of available training data between new classes and the stored classes may still vary a lot during the training phase due to the limited memory budget. Existing work~\cite{liu2022long} addresses this problem by decoupling the training process into two stages to first learn a feature extractor and then fine-tune the classifier using a class-balanced sampler. In this work, we propose to use Balanced Softmax (BS)~\cite{BSLoss} by extending it into a long-tailed continual learning scenario without requiring a decoupled training process. Specifically, during the training phase of the new task $\mathcal{T}^{i}$, a distribution vector $v_d \in \mathbb{R}^{C_{1:i}} $ is generated by counting the number of training data of each food class for input images in the current task. Recall $L \in \mathbb{R}^{C_{1:i}}$ is the output logits from current model $h_i$, the distribution vector is then used as the prior information when calculating the loss as shown in (\ref{eq: bsloss})
\begin{equation}
    \label{eq: bsloss}
    \mathcal{L}_{bs} =  \sum_{k=1}^{C_{1:i}} -y^{k}log[\Phi(\bar{v}_d^k+L^k)]
\end{equation}
where $\bar{v}_d = v_d / sum(v_d)$ is the normalized distribution vector and $\Phi()$ denotes the $Softmax$ function. Therefore, the larger value in the distribution vector achieves smaller gradients when we compute the cross-entropy using the adjusted logits $v_t+L$ and vice versa. This addresses the class-imbalance issue and enables the end-to-end training pipeline. 

The overall training loss function is the weighted sum of feature-based knowledge distillation $\mathbf{}$ as described in (\ref{eq: fkd}) and the balanced softmax $\mathcal{L}_{bs}$, which can be expressed as 
\begin{equation}
    \label{eq: finalloss}
    \mathcal{L} = \mathcal{L}_{bs} + \lambda \mathcal{L}_{fkd}
\end{equation}
where $\lambda$ is the adaptive ratio to tune the two losses. In this work, as the number of training data $D_i$ may vary a lot for each task $\mathcal{T}^{i}$, we propose to calculate $\lambda = \sqrt{ \vert D_i\vert / \vert D_{1:i}\vert }$ as the ratio of training data for the current task and the learned tasks. Therefore, the ratio $\lambda$ increases when there are more training data from new classes.

\begin{table*}[t]
    \caption{Results on Food101-LT, VFN-LT, VFN186-LT, VFN186-INSULIN, VFN186-T2D, and ImageNetSubset-LT by comparing with existing continual learning methods in terms of average accuracy $A_M$ ($\%$). Best results are marked in bold.}
    \label{tab:compare_sota_merged}
    \centering
    \renewcommand{\arraystretch}{1.2} 
    \scalebox{0.9}{
    \begin{tabular}{>{\centering\arraybackslash}m{2.6cm}|>{\centering\arraybackslash}m{1.4cm}|>{\centering\arraybackslash}m{1.4cm}|>{\centering\arraybackslash}m{1.4cm}|>{\centering\arraybackslash}m{1.4cm}|>{\centering\arraybackslash}m{1.4cm}|>{\centering\arraybackslash}m{1.4cm}|>{\centering\arraybackslash}m{1.4cm}|>
    {\centering\arraybackslash}m{1.4cm}|>{\centering\arraybackslash}m{1.4cm}}

        \hline
        \textbf{Datasets} & \multicolumn{3}{c|}{\textbf{Food101-LT}} & \textbf{VFN-LT} & \textbf{VFN186-LT} & \textbf{VFN186-INSULIN} & \textbf{VFN186-T2D} & \multicolumn{2}{c}{\textbf{ImageNetSubset-LT}}  \\
        \hline
        \textbf{Number of tasks} & $N=5$ & $N=10$ & $N=20$ & $N=7$ & $N=9$ & $N=9$ & $N=9$ & $N=10$ & $N=20$ \\
        \hline
        LwF~\cite{LWF} & 10.02 & 5.86 & 0.83 & 4.85 & 11.60 & 10.99 & 11.12 & 23.55  & 20.16 \\
        EWC~\cite{kirkpatrick2017overcoming} & 5.05 & 3.70 & 0.83 & 6.31 & 11.09 & 10.40 & 4.12 & 19.73  & 16.49 \\
        iCaRL~\cite{ICARL} & 12.42 & 12.46 & 11.04 & 18.76 & 8.76 & 7.23 & 7.56 & 33.75  & 29.71\\
        LwM~\cite{dhar2019lwm} & 10.82 & 7.22 & 2.45 & 12.32 & 11.04 & 10.37 & 10.88 & 30.24  & 26.63 \\
        IL2M~\cite{dualmemory} & 11.45 & 10.97 & 6.81 & 18.68 & 11.23 & 10.52 & 11.30 & 31.70  & 25.20  \\
        BiC~\cite{BiC} & 16.72 & 12.39 & 10.38 & 20.89 & 9.09 & 9.27 & 11.08 & 33.31  &  30.86 \\
        EEIL-2stage~\cite{liu2022long} & 14.96 & 13.29 & 9.76 & 22.98 & 12.86 & 11.74 & 12.69 & 36.84  & 30.39 \\
        LUCIR-2stage~\cite{liu2022long} & 18.90 & 13.03 & 10.85 & 24.26 & 15.80 & 13.64 & 14.88 & 39.87 &  34.79  \\
        PODNet-2stage~\cite{liu2022long} & 17.89 & 11.12 & 10.28 & 25.58 & 16.00 & 13.77 & 15.11 &  34.79  & 31.71 \\
        MAFDRC~\cite{chen2023dynamic} & 19.04 & 16.20 & 13.63 & 22.48  & 18.59 &  17.03 & 18.54 & 40.01 & 34.48  \\
        DGR~\cite{He_2024_CVPR} &  23.08 & 20.35 & 16.43 & 26.11 & 19.58 & 17.66 & 17.90 & \textbf{45.12}  & \textbf{40.79} \\
        \hline
        \textbf{Ours} & \textbf{27.52} & \textbf{25.12} & \textbf{21.72} & \textbf{27.53} & \textbf{22.21} & \textbf{20.61} & \textbf{21.23}  & 44.68  & 40.31 \\
        \hline
    \end{tabular}
    }
\end{table*}

\vspace{-0.3em}
\section{Experiment}
\label{sec:experiment}
In this section, we evaluate our proposed long-tailed continual learning framework for visual food recognition as illustrated in Section~\ref{sec: method}. Specifically, we first introduce the experimental setup including the split of datasets and implementation detail described in Section~\ref{subsec: datasets for exp} and~\ref{subsec: implementation detail}. Then we compare our method with existing work in Section~\ref{subsec: compare sota} and conduct an ablation study to show the effectiveness of each individual component in Section~\ref{subsec: ablation study}. Finally, we discuss potential techniques that can boost the performance of real-world food-related applications in Section~\ref{subsec: discussion}. 

\vspace{-0.5em}
\subsection{Datasets}
\label{subsec: datasets for exp}

\textbf{Food101-LT} is the long-tailed version of Food-101~\cite{Food-101}, created using the \textit{Pareto distribution}~\cite{pareto_dist} with the power ratio of $\alpha = 6$. We randomly partition the 101 food classes into 5, 10, and 20 tasks for continual learning, where each task introduces 20, 10, and 5 new classes, respectively, except the first task with one extra class. The test set is kept as balanced with 125 images per class. 

\textbf{VFN-LT} is a long-tailed version of VFN~\cite{mao2020visual} based on food consumption frequency of healthy people.  The 74 food classes are split into 7 tasks, with the first task containing 14 new classes and the remaining tasks containing 10 new classes. The test set has 25 images per class.

\textbf{VFN186-LT, VFN186-INSULIN} and \textbf{VFN186-T2D} are long-tailed versions of VFN186. VFN186-LT is created similarly to VFN-LT, while VFN186-INSULIN and VFN186-T2D are based on food consumption frequencies of insulin takers and individuals with type 2 diabetes without insulin. We divide 186 food classes into $N=9$ tasks with the first task containing 26 new classes and the rest containing 20. To facilitate an equatable analysis, we use the same testing data in VFN186-LT, VFN186-INSULIN and VFN186-T2D, which is balanced with 25 samples per class, totaling 4,650 images. 

\textbf{ImageNetSubset-LT} is a subset of ImageNet~\cite{russakovsky2015imagenet}. We follow~\cite{liu2022long, He_2024_CVPR} to select 100 classes and apply the long-tailed transformation. Specifically, we randomly remove training samples following an imbalance factor of $\rho = n_{\text{max}}/n_{\text{min}} = 100$, where $n_{\text{max}}$ and $n_{\text{min}}$ denote the maximum and minimum number of training samples per class, respectively. The dataset is split evenly into $N=10$ task and $N=20$ tasks, where each task contains 10 and 5 new classes, respectively. The test set remains unchanged, preserving the original class-balanced distribution.

\vspace{-1em}
\subsection{Implementation Detail}
\label{subsec: implementation detail}
Our implementation of neural networks are based on the Pytorch framework and we apply the ResNet-18 from scratch as the backbone for all experiments. The ResNet implementation follows the setting suggested in~\cite{RESNET}. We train each new task for 90 epochs with the learning rate starting from 0.1 and decreasing with a ratio of $1/10$ for every 30 epochs. The batch size is set to 128 and we apply the stochastic gradient descent (SGD) optimizer with a weight decay of $0.0001$. To ensure a fair comparison between our method and existing methods, we set the random seed to 1993, following~\cite{ICARL,luo2023class}.

\textbf{Exemplar Selection Strategy: }To construct the memory buffer, we follow the benchmark setting in~\cite{He_2024_CVPR} and set the memory budget to $\mathcal{M} = 20$, allowing the storage of at most 20 exemplars per class. We adopt herding algorithm~\cite{HERDING} for exemplar selection, which selects samples closest to the class mean in the feature space as exemplars. For tailed classes are with fewer than $\mathcal{M}$ available samples, we employ our CAM-based exemplar augmentation strategy to generate synthetic exemplars, ensuring balanced memory utilization across all classes.

\textbf{Evaluation protocol:} We use Top-1 classification accuracy to evaluate the model after each task $\mathcal{T}^{i}$ on test data covering previously seen classes $C_{1:i}$. Besides, we report the average accuracy $A_{M}$, calculated by averaging the accuracy after each task, which shows the overall performance across the continual learning procedure. Each experiment is run five times and the average performance is presented.

\begin{figure*}[t]
\begin{center}
  \includegraphics[width=1.\linewidth]{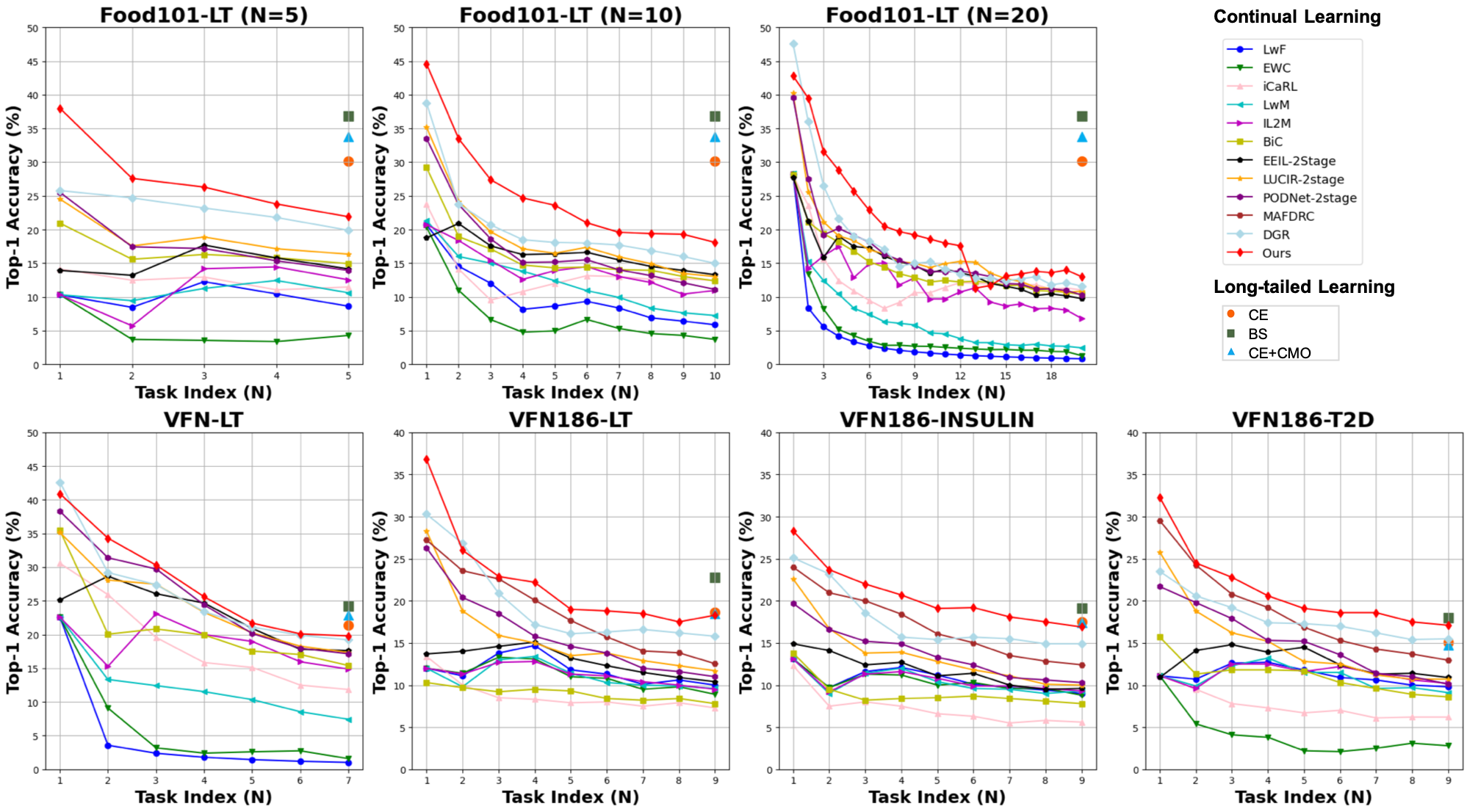}
  \caption{Results on Food101-LT, VFN-LT, VFN186-LT, VFN186-INSULIN and VFN186-T2D with different number of tasks $N$. Each marker represents the Top-1 classification accuracy evaluated on all classes seen so far after learning each task.}
  \label{fig:results-sota}
\end{center}
\vspace{-0.5cm}
\end{figure*}

\vspace{-0.5em}
\subsection{Comparisons With Existing Methods}
\label{subsec: compare sota}

\subsubsection{Performance across Datasets}
 Table~\ref{tab:compare_sota_merged} summarizes the average accuracy $A_M$ on Food101-LT, VFN-LT, VFN186-LT, VFN186-INSULIN, VFN186-T2D, and ImageNetSubset-LT. 
 Our method shows significant improvements, particularly on Food101, with different numbers of tasks $N \in \{5, 10, 20\}$, achieving approximately a 5\% increase in accuracy. Moreover, there are enhancements on VFN186-LT, VFN186-INSULIN, and VFN186-T2D, which feature more imbalanced distributions as discussed in Section~\ref{sec: dataset}. For example, on three long-tailed VFN186, we achieve about a $7\%$ increase over the 2-stage framework even without requiring a decoupled training process and a $3\%$ improvement compared to DGR. However, tends often decreases as the total number of tasks $N$ increases. Therefore, we need to address the catastrophic forgetting to maintain the learned knowledge at each learning phase of new tasks after the first task. 
However, improvements are not evident on VFN. Considering the imbalanced data and the differences between food images are much smaller than those in other scenarios, this type of tasks is more difficult, making it hard to achieve significant improvements in classification results. Additionally, we do not present average performance in general scenarios since the random seed is fixed. In this case, we only evaluate the algorithm itself, rather than its performance across various situations. 
To demonstrate the broader applicability of our method beyond food data, we conduct additional experiments on ImageNetSubset for long-tailed learning across diverse object categories. Despite being primarily designed for food-related tasks, our approach shows strong generalization capability, achieving competitive performance against state-of-the-art methods. These results underscore the robustness of our method to effectively handle class imbalance, suggesting the potential for visual recognition tasks across multiple domains.

\subsubsection{Visualization and Upper Bound}
Figure~\ref{fig:results-sota} shows top-1 classification accuracy across all seen classes after each task and upper bounds for continual learning.
We use the joint training results as the upper bounds. Specifically, we train models on the full long-tailed datasets using three different strategies: vanilla cross-entropy (CE), balanced softmax (BS)\cite{ren2020balanced}, and Context-rich Minority Oversampling (CMO)\cite{park2021cmo}. To ensure a fair comparison with the long-tailed continual learning experiments, the backbone and configurations are keep the same of other experiments. We report the final performance on the last task, as joint training is equivalent to training on the entire dataset at once. As shown in Figure~\ref{fig:results-sota}, these three settings achieve the best overall performance on Food101-LT and VFN-LT. Notably, on VFN186, CE and CE+CMO even perform worse than continual learning methods. This is due to the severe class imbalance and the limited number of samples in certain categories. Notably, as discussed in \cite{nu15122751}, the sample counts of rare classes in VFN186 are significantly smaller than those in Food101-LT and VFN-LT, which exacerbates the performance drop on VFN186.

 Our method achieves promising performance at each stage of the new task. 
 Interestingly, in long-tailed scenarios, unlike conventional continual learning where accuracy typically declines over time, we observe improvements after learning new tasks sometimes. For example, on VFE186-LT, accuracy increases for LwF, BiC, LUCIR, PODNet, and EEIL-2Stage after the task $N=4$. This occurs because the number of training samples varies significantly among different tasks in long-tailed continual learning where the model gains better knowledge for tasks with a larger number of training images. Therefore, it is common for accuracy to improve when learning classes with larger sample sizes in long-tailed continual learning. However, it also imposes new challenges in handling class-imbalance across different tasks and hyper-parameter tuning (\textit{e.g. }the knowledge distillation factor in Equation~\ref{eq: og_crosskd}).

\subsubsection{Comparisons With Prompt-Based Continual Learning}
We further compare our method with recent prompt-based continual learning approaches including L2P~\cite{DBLP:journals/corr/abs-2112-08654}, DualPrompt~\cite{wang2022dualpromptcomplementarypromptingrehearsalfree}, and CODA-Prompt~\cite{smith2023codapromptcontinualdecomposedattentionbased}. We adopt ViT-B/16~\cite{vit} pretrained on ImageNet-21K~\cite{imagenet21k} as our backbone architecture across all experiments. We train each new task for $20$ epochs with batch size of $48$ using SGD optimizer. The initial learning rate is set to $0.05$ and decays following a cosine scheduler. To ensure fair comparisons, we follow~\cite{DBLP:journals/corr/abs-2112-08654} to select the same amount of exemplars as used in our method. Each experiment is conducted five times, and the reported results represent the mean accuracy with standard deviation on Food101-LT for task numbers $N=10$ and $N=20$. As shown in Table~\ref{tab:prompt}, our method consistently outperforms all prompt-based methods on Food101-LT for all tasks, and the relatively low standard deviation in our experimental results indicates that our method provides stable performance.  These results indicate our strong adaptation ability to work with pre-trained backbones.

\begin{table}[t]
    \caption{Results on Food101-LT by comparing with recent prompt-based incremental learning methods in terms of average accuracy ($\%$) $\pm$ STD. 
    }
    \label{tab:prompt}
    \centering
    \scalebox{1.0}{
\begin{tabular}{c|cc}
\hline
                 & \multicolumn{2}{c}{Food101-LT} \\
                 &  $N=10$         & $N=20$          \\ \hline
L2P-R~\cite{DBLP:journals/corr/abs-2112-08654}            &    $82.01 \pm 0.89$       &   $75.58 \pm 0.52$       \\
DualPrompt~\cite{wang2022dualpromptcomplementarypromptingrehearsalfree}       &   $83.33 \pm 0.25$  &  $74.29 \pm 1.41$    \\
CODA-Prompt~\cite{smith2023codapromptcontinualdecomposedattentionbased}      &   $85.14 \pm 0.39$   &   $78.72 \pm 0.41$ \\
Ours             &   \textbf{$86.52 \pm 0.18$}   &   \textbf{$80.45 \pm 0.48$}      \\ \hline
\end{tabular}
}
\vspace{-0.4cm}
\end{table}

\begin{table}[t]
     \caption{Ablation study on Food101-LT and VFN-LT in terms of average accuracy $A_M$.  }
    \label{tab:ablation}
    \centering
    \scalebox{0.9}{
    \begin{tabular}{ccc|cccc}
        \hline
        & & & \multicolumn{3}{c}\textbf{Food101-LT \quad \quad \quad \quad} & VFN-LT \\
        $\bm{\mathcal{L}_{fkd}}$ & \textbf{CAM-CutMix} & $\bm{\mathcal{L}_{bs}}$ & $N=5$ & $N=10$ & $N=20$ & $N=7$ \\
         \hline
         & & & 5.90 & 8.79 & 10.55 & 12.21 \\
         \checkmark  & & & 17.42 & 15.83 & 13.96 & 22.53 \\
         & \checkmark & & 13.27 & 12.99 & 11.64 & 16.73 \\
         & & \checkmark & 16.52 & 14.20 & 12.02 & 22.96 \\
         \checkmark & \checkmark & & 19.31 & 17.26 & 15.49 & 24.18 \\
         \checkmark  &\checkmark &\checkmark & \textbf{21.83} & \textbf{19.25} & \textbf{17.43} & \textbf{29.33} \\
        \hline
    \end{tabular}
    }
\end{table}

\begin{figure}[t]
\begin{center}
  \includegraphics[width=1.\linewidth]{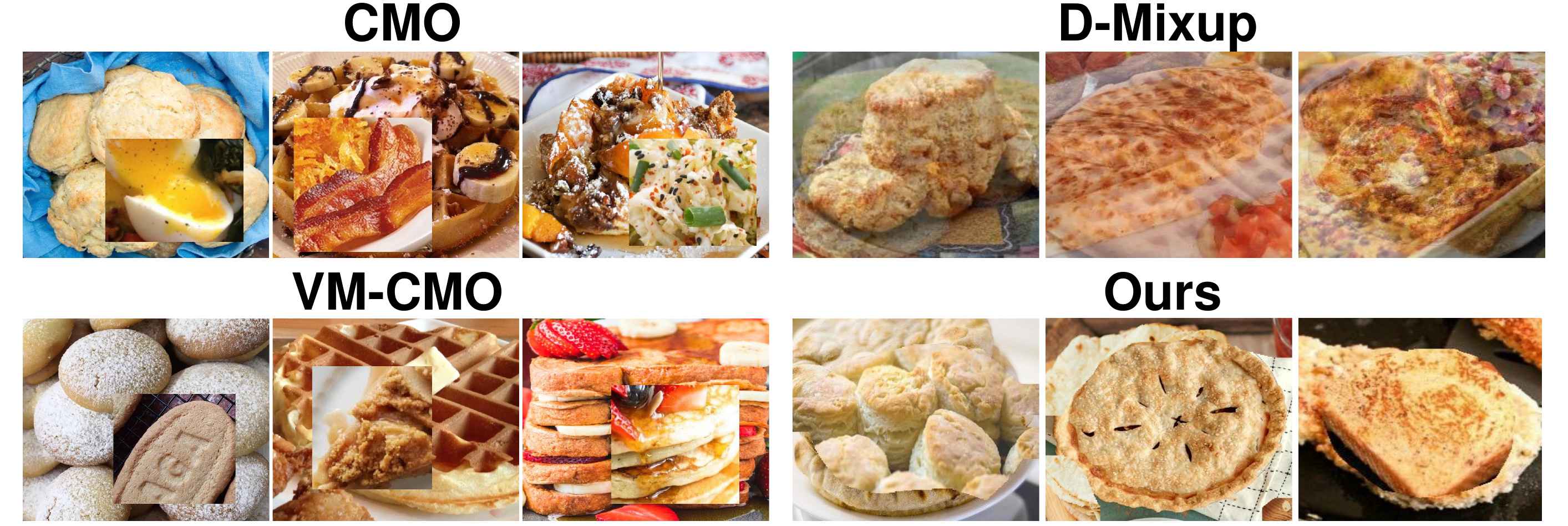}
  \caption{Examples of augmented food images on VFN-LT using CMO~\cite{CMO}, VM-CMO~\cite{he2022long}, D-Mixup~\cite{gao2022dynamic_LTingredient} and our proposed CAM-CutMix. }
  \label{fig:augmentation}
  \vspace{-0.5cm}
\end{center}
\end{figure}

\vspace{-0.8em}
\subsection{Ablation Study}
\label{subsec: ablation study}

\subsubsection{Effectiveness of Core Components}
We evaluate the effectiveness of each individual component in our proposed framework including (i) the feature-based knowledge distillation ($\bm{\mathcal{L}_{fkd}}$), (ii) the cam-based data augmentation (\textbf{CAM-CutMix}) and (iii) the integration of balanced softmax with adaptive ratio ($\bm{\mathcal{L}_{bs}}$). Formally, we consider the \textit{baseline} method as using an imbalanced memory buffer ($\mathcal{M} = 20$) with cross-entropy loss and integrating each of the components mentioned above to conduct experiments. The results in terms of average accuracy $A_M$ are summarized in Table~\ref{tab:ablation}. We observe consistent performance improvements compared with \textit{baseline} by adding our proposed techniques. Specifically, the feature-based knowledge distillation $\bm{\mathcal{L}_{fkd}}$ achieves the largest improvements on the Food101-LT dataset, demonstrating that catastrophic forgetting is a crucial issue and the integration with CAM-CutMix can achieve higher accuracy. On the other hand, as VFN-LT exhibits more severe class-imbalance problems due to a higher imbalance ratio, the balanced softmax $\bm{\mathcal{L}_{bs}}$ term has the most significant impact, resulting in the largest performance improvements. 
By integrating all three components, our proposed framework obtains the best classification accuracy on these two datasets.

\subsubsection{Effectiveness of CAM-CutMix}
We further evaluate our proposed CAM-CutMix by replacing it with existing data augmentation based methods including the CutMix~\cite{yun2019cutmix} based approaches: (a) the original CutMix used in \textbf{CMO}~\cite{CMO}, (b) Visual-Multi CutMix (\textbf{VM-CMO})~\cite{he2022long}, (c) \textbf{SnapMix}~\cite{huang2021snapmix} and the Mixup~\cite{zhang2018mixup} based approach: (d) \textbf{D-Mixup}~\cite{gao2022dynamic_LTingredient}. We conduct experiments on VFN-LT and Food101-LT with $N=10$ as shown in Table~\ref{tab:augmentation}. Generally, the CutMix-based approaches work better in long-tailed continual learning scenarios than D-Mixup, which is usually applied in multi-label recognition scenarios. In addition, the SnapMix achieves a slightly better performance than CMO and VM-CMO as it also considers the class-activation map (CAM) when generating mixed labels. Our method achieves the best performance as it not only preserves the most important regions based on CAM but also enables seamless CutMix, rather than relying on a randomly generated bounding box. The example augmented food images using VFN-LT are shown in Figure~\ref{fig:augmentation}. Note that we do not visualize SnapMix~\cite{huang2021snapmix} as it has the same synthetic image as in CMO~\cite{CMO} but with a different mixed label. 

\begin{table}[t]
    \centering
     \caption{Ablation study of different data augmentation methods on Food101-LT ($N=10$) and VFN-LT with average accuracy $A_M$.  }
    \label{tab:augmentation}
    \scalebox{1}{
    \begin{tabular}{ccc}
        \hline
         & Food101-LT ($N=10$) & VFN-LT \\
         \hline
         CMO~\cite{CMO} & 17.28 & 25.93 \\ 
         VM-CMO~\cite{he2022long} & 16.47 & 26.41 \\ 
         SnapMix~\cite{huang2021snapmix} & 18.31 & 27.62 \\ 
         D-Mixup~\cite{gao2022dynamic_LTingredient} & 15.93 & 25.14 \\ 
         CAM-CutMix (Ours) & \textbf{19.25} & \textbf{29.33} \\          
        \hline
    \end{tabular}
    }
\end{table}

\subsubsection{Robustness to Design Variations}
To evaluate the robustness of our method, we conducte additional experiments by replacing key components with alternative approaches: (i) we substitute our herding exemplar selection with random selection (denoted as Random) for memory replay, (ii) we replace the cosine embedding loss with Mean Squared Error (MSE) loss in the knowledge distillation module, and (iii) we use Grad-CAM instead of the original CAM approach for generating attention maps. The results on Food101-LT with different task numbers ($N=5, 10, 20$) are shown in Table~\ref{tab:varient}, where the up arrow $\uparrow$ indicates an improvement over the original setting while the down arrow $\downarrow$ denotes a performance drop. The variations in performance across different configurations highlight the flexibility of our method, as it maintains stable accuracy regardless of the specific component used.  
Grad-CAM augmentation yields the most consistent improvements (e.g., $\uparrow 2.27$ at $N=10$), demonstrating its effectiveness in selecting semantic important regions. MSE loss results in mixed performance, with minor improvements at $N=10$ but drops at $N=5$ and $N=20$. Overall, these results confirm that our framework is adaptable and can accommodate different design choices while maintaining strong continual learning performance.

\begin{table}[t]
    \caption{Ablation study on Food101-LT with different variants. Results are reported as average accuracy (\%) across different task settings ($N=5, 10, 20$).}
    \label{tab:varient}
    \vspace{-0.3em}
    \centering
    \renewcommand{\arraystretch}{1.2} 
    \begin{tabular}{l|ccc}
        \hline
         & \multicolumn{3}{c}{\textbf{Food101-LT}} \\
         & \textbf{$N=5$} & \textbf{$N=10$} & \textbf{$N=20$} \\
        \hline
        Ours w/ Random  & 27.74\textsubscript{$(\uparrow 0.22)$}  & 24.88\textsubscript{$(\downarrow 0.24)$} & 21.53\textsubscript{$(\downarrow 0.19)$} \\
        Ours w/ MSE Loss       & 26.75\textsubscript{$(\downarrow 0.77)$}  & 25.38\textsubscript{$(\uparrow 0.26)$}   & 21.25\textsubscript{$(\downarrow 0.47)$}  \\
        Ours w/ Grad-CAM      & 28.83\textsubscript{$(\uparrow 1.31)$}   & 27.39\textsubscript{$(\uparrow 2.27)$}   & 23.08\textsubscript{$(\uparrow 1.36)$}  \\
        Ours (original)               & 27.52   & 25.12  & 21.72  \\
        \hline
    \end{tabular}
\end{table}

\vspace{-0.8em}
\subsection{Discussions}
\label{subsec: discussion}
Despite the performance improvements our framework demonstrates compared to existing methods as shown in Table~\ref{tab:compare_sota_merged}, the deployment in real-world applications remains challenging due to current classification accuracy and computational complexity. Therefore, in this section, we (1) analyze the running time of different methods to assess computational efficiency, and discuss potential techniques that could be applied to boost the performance including (2) increasing the memory buffer capacity to store more exemplar images for knowledge replay and (3) performing transfer learning on our methods to evaluate its scalability across different pretraining.

\subsubsection{Computational Complexity}
\begin{figure}[t]
    \begin{center}
   \includegraphics[width=1.\linewidth]{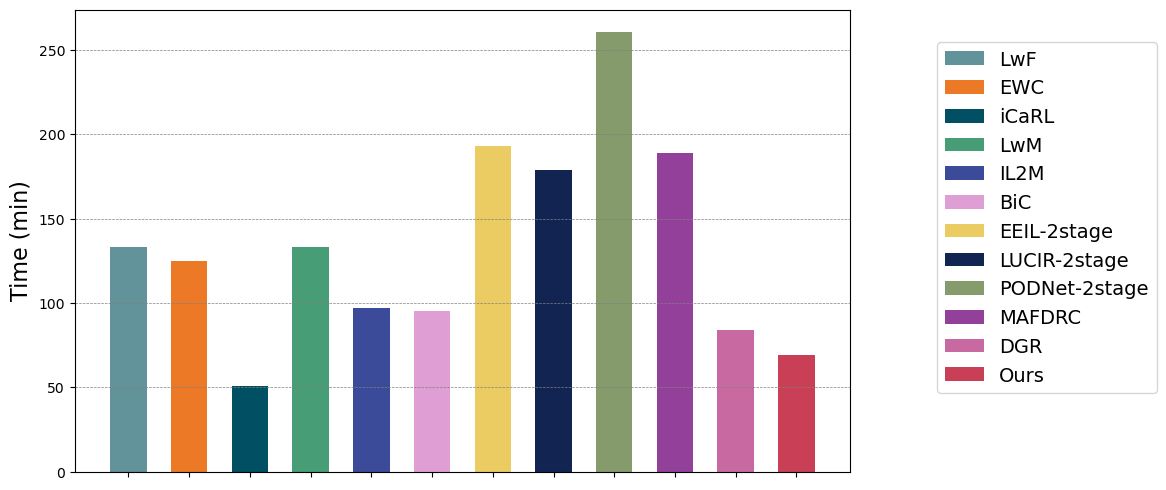}
   
  \caption{Running time (min) comparison on VFN186-LT for different models.}
  \label{fig:running}
    \end{center}
    \vspace{-0.8cm}
\end{figure}

Regarding computational complexity, we record the time required for each model to be trained from scratch, including the time needed for testing. Our method takes 69 minutes to finish the whole training process. As seen in Figure~\ref{fig:running}, there is a 15-minute reduction compared to the recently released DGR~\cite{He_2024_CVPR} method, which requires 84 minutes, but it achieves superior classification performance, making it both time-efficient and effective. iCaRL~\cite{ICARL} stands out as the fastest but at the cost of lower classification accuracy, particularly in handling datasets of long-tailed distribution. Three two-stage methods show robust classification accuracy but at the expense of significantly higher training times, which are all over three times to our approach's. Our method strikes an optimal balance between computational efficiency and classification accuracy, outperforming all other methods when considering both aspects, which makes it particularly suitable for real-world applications.

\subsubsection{Memory buffer capacity}
\label{subsubsec: memory capacity}
As one of the most efficient techniques to address catastrophic forgetting, the performance of knowledge replay greatly relies on the capacity of the memory buffer (\textit{i.e. }how many exemplar images can be stored). In this part, we evaluate the long-tailed continual learning performance by varying the memory buffer capacity $\mathcal{M} \in \{10, 20, 30, 40, 50, 100\}$. To ensure fair class-wise representation under the long-tailed setting, we adopt a fixed number of exemplars per class. Table~\ref{tab:explist} reports the average accuracy $A_M$ on Food101-LT ($N=10$) and VFN-LT.
We observe consistent performance improvements as $\mathcal{M}$ increases. However, the memory buffer capacity is a significant constraint for continual learning in real-world applications as it requires larger memory storage and also poses challenges related to privacy concerns when storing original images as exemplars. Additionally, the gain saturates at different points depending on the dataset. For instance, Food101-LT continues to improve up to $\mathcal{M}=40$, while VFN-LT shows marginal improvement beyond $\mathcal{M}=20$. This suggests a dataset-dependent trade-off between buffer size and performance. Moreover, the performance bottleneck is predominantly due to dual challenges of catastrophic forgetting and class-imbalance problems that arise in the long-tailed continual learning scenario. 


\begin{table}[htbp]
    \centering
    \caption{Average accuracy ($A_M$) on Food101-LT ($N=10$) and VFN-LT by varying the memory buffer capacity $\mathcal{M} \in \{10, 20, 30, 40, 50, 100\}$.}
    \label{tab:explist}
    \begin{tabular}{c|cccccc}
        \hline
        \textbf{Buffer Size} & \textbf{10} & \textbf{20} & \textbf{30} & \textbf{40} & \textbf{50} & \textbf{100} \\
        \hline
        \textbf{Food101-LT} & 16.96 & 19.69 & 20.71 & 22.15 & 23.14 & 24.30 \\
        \textbf{VFN-LT} & 25.71 & 29.33 & 30.84 & 31.25 & 31.89 & 32.55 \\
        \hline
    \end{tabular}
\end{table}

\subsubsection{Variants of Backbones and Pre-training Datasets}
\label{subsubsec: transfer learning}
 Applying the deep models pre-trained on large-scale image datasets as the backbone is a common strategy to enhance performance in many vision tasks~\cite{hendrycks19a_pretrain, Food2K}. In this part, we investigate how our method performs when applied to different backbone architectures and pre-training datasets. Instead of modifying the learning paradigm, we analyze whether our approach remains effective across various network structures and different levels of pre-training.
 We consider backbones with various depth including \textit{ResNet-50}~\cite{RESNET}, \textit{MobileNet}~\cite{howard2017mobilenets}, \textit{EfficientNet}~\cite{tan2019efficientnet} \textit{Vision Transformers (ViT)}~\cite{vit} and its variants \textit{DeiT}~\cite{deit-vit} and \textit{Swin}~\cite{swin-vit} transformers. In addition, we leverage ImageNet-1K~\cite{IMAGENET1000} and ImageNet-21K~\cite{imagenet21k} as the pre-training datasets. ImageNet-1K contains 1,000 classes of general objects, which is the subset of full ImageNet-21K that contains 21,841 classes with over 14,197,122 training images. The VFN-LT results in average accuracy $A_M$ are shown in Table~\ref{tab:models}. We observe over $20\%$ performance improvements by using pre-trained models on large-scale datasets compared to our results in Table~\ref{tab:compare_sota_merged} with a model from scratch. It manifests that pre-training enhances the backbone network's feature extraction capabilities, thereby yielding the most discriminative features essential for downstream tasks. In addition, pre-training on larger-scale datasets with more images and classes makes higher accuracy. However, there is a trade-off between the computation complexity and the performance where the increase of model parameters would require longer training time and higher computation capability, which may not be practical for specific real-world applications with limited resources. Note that we intentionally refrain from utilizing food datasets for pre-training in this part to prevent potential overlap with any food class in VFN~\cite{mao2020visual}, though there may be a more substantial performance enhancement if pre-trained on large-scale food datasets such as Food2K~\cite{Food2K}.

\vspace{-1.0em}

\begin{table}[htbp]
    \centering
    \caption{Average accuracy ($A_M$) on VFN-LT by leveraging pre-trained models.}
    \label{tab:models}
    \scalebox{0.82}{
    \begin{tabular}{c|cccccc}
        \hline
        \textbf{Model} & \textbf{MobileNet} & \textbf{ResNet} & \textbf{EfficientNet} & \textbf{ViT} & \textbf{DeiT} & \textbf{Swin} \\
        \textbf{Parameters (10M)} & 1.8 & 2.5 & 5.4 & 8.2 & 8.5 & 8.8 \\
        \hline
        \textbf{ImageNet-1K} & 54.99 & 56.73 & 61.78 & 59.75 & 61.41 & 63.17 \\
        \textbf{ImageNet-21K} & 56.64 & 58.92 & 63.83 & 64.79 & 65.01 & 71.18 \\
        \hline
    \end{tabular}
    }
\end{table}

\vspace{-1.0em}

\section{Conclusion}
\label{sec: conclusion}

In this work, we focus on visual food recognition in long-tailed continual learning. We create an expanded dataset VFN186 and its three benchmark long-tailed food image datasets that exhibit the real-life food consumption frequency. The proposed end-to-end framework combines effective feature-based knowledge distillation and a novel data augmentation module, capable of learning new food classes in long-tailed data distribution without forgetting the learned knowledge. Our method outperforms existing approaches on all mentioned datasets. Future work includes developing an exemplar-free framework to tackle issues related to large memory buffers and privacy concerns with stored food images.

\bibliographystyle{IEEEbib}
{\small \bibliography{refs}}

\newpage

\vspace{11pt}

\vfill

\end{document}